\documentclass[sigconf]{acmart}
\usepackage{microtype}      
\usepackage{xcolor}         
\usepackage{diagbox}
\usepackage{svg}
\usepackage{graphicx}

\usepackage{multicol}
\usepackage{multirow}

\usepackage{makecell}

\usepackage{amsmath}

\usepackage{caption}
\usepackage{subcaption}

\usepackage[ruled,linesnumbered,vlined]{algorithm2e}
\usepackage{algpseudocode}
\usepackage{multirow}

\AtBeginDocument{%
    }

\copyrightyear{2023}
\acmYear{2023}
\setcopyright{rightsretained}
\acmConference[MM '23]{Proceedings of the 31st ACM International Conference on Multimedia}{October 29-November 3, 2023}{Ottawa, ON, Canada}
\acmBooktitle{Proceedings of the 31st ACM International Conference on Multimedia (MM '23), October 29-November 3, 2023, Ottawa, ON, Canada}
\acmDOI{10.1145/3581783.3612032}
\acmISBN{979-8-4007-0108-5/23/10}

\makeatletter
\gdef\@copyrightpermission{
\begin{minipage}{0.3\columnwidth}
 \href{https://creativecommons.org/licenses/by/4.0/}{\includegraphics[width=0.90\textwidth]{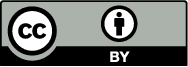}}
\end{minipage}\hfill
\begin{minipage}{0.7\columnwidth}
 \href{https://creativecommons.org/licenses/by/4.0/}{This work is licensed under a Creative Commons Attribution International 4.0 License.}
\end{minipage}
\vspace{5pt}
}
\makeatother

\usepackage{xspace}
\newcommand{\sys}{PatchBackdoor\xspace}
\newcommand{\eg}{{\it e.g.,}\xspace}
\newcommand{\ie}{{\it i.e.,}\xspace}

\begin{document}

\title{PatchBackdoor: Backdoor Attack against Deep Neural Networks without Model Modification}

\author{Yizhen Yuan}
\affiliation{
    \institution{Institute for AI Industry Research (AIR), Tsinghua University}
    \city{Beijing}
    \country{China}
}
\email{yuanyz21@mails.tsinghua.edu.cn}

\author{Rui Kong}
\affiliation{
    \institution{Shanghai Jiao Tong University}
    \city{Shanghai}
    \country{China}
}
\email{kongrui@sjtu.edu.cn}

\author{Shenghao Xie}
\affiliation{
    \institution{Wuhan University}
    \city{Wuhan}
    \country{China}
}
\email{xieshenghao@whu.edu.cn}

\author{Yuanchun Li}
\authornote{Corresponding author.}
\affiliation{
    \institution{Institute for AI Industry Research (AIR), Tsinghua University
    }
    \city{Beijing}
    \country{China}
}
\affiliation{
    \institution{Shanghai AI Laboratory}
    \city{Shanghai}
    \country{China}
}
\email{liyuanchun@air.tsinghua.edu.cn}

\author{Yunxin Liu}
\affiliation{
    \institution{Institute for AI Industry Research (AIR), Tsinghua University
    }
    \city{Beijing}
    \country{China}
}
\affiliation{
    \institution{Shanghai AI Laboratory}
    \city{Shanghai}
    \country{China}
}
\email{liuyunxin@air.tsinghua.edu.cn}

\begin{abstract}
Backdoor attack is a major threat to deep learning systems in safety-critical scenarios, which aims to trigger misbehavior of neural network models under attacker-controlled conditions.
However, most backdoor attacks have to modify the neural network models through training with poisoned data and/or direct model editing, which leads to a common but false belief that backdoor attack can be easily avoided by properly protecting the model.
In this paper, we show that backdoor attacks can be achieved without any model modification. Instead of injecting backdoor logic into the training data or the model, we propose to place a carefully-designed patch (namely backdoor patch) in front of the camera, which is fed into the model together with the input images. The patch can be trained to behave normally at most of the time, while producing wrong prediction when the input image contains an attacker-controlled trigger object.
Our main techniques include an effective training method to generate the backdoor patch and a digital-physical transformation modeling method to enhance the feasibility of the patch in real deployments.
Extensive experiments show that \sys can be applied to common deep learning models (VGG, MobileNet, ResNet) with an attack success rate of 93\% to 99\% on classification tasks.
Moreover, we implement \sys in real-world scenarios and show that the attack is still threatening.
\end{abstract}

\maketitle

\section{Introduction}

Deep Neural Networks (DNNs) are widely used in many security-critical edge systems such as autonomous driving \cite{autodriving}, face authentication \cite{face} and medical diagnosis \cite{medicalEye,medicalReasoning}. While bringing great convenience in many applications, the security issues of deep learning (DL) are also gaining extensive attention.

It is widely known that DNN is vulnerable to many types of attacks, and the backdoor attack is a major one of them. Most backdoor attack approaches conduct the attack by training the victim model with poisoned datasets \cite{badnet,smallBackdoor}. The trained model will have a high benign accuracy when normal test samples are predicted, while the model will give wrong predictions when certain attacker-controlled triggers are present. Some other attackers conduct the attack by directly modifying the model structures and/or weights\cite{modelWeightModify}, which usually happens in third-party machine learning platforms where users outsource the training or serving to untrusted service providers. The attackers can modify their models to inject backdoors before the models are actually deployed.

A primary limitation of the backdoor attack is the need to modify the model, which could be challenging in most security-critical scenarios. For instance, most autonomous driving companies use self-collected and carefully-filtered datasets for training and will not outsource the training to cloud service either. When being deployed, the model can be placed in the read-only memory to ensure integrity. Therefore, although the backdoor attack seems threatening, it is less concerning for most model developers that can securely manage the training datasets and deployed models.

In this paper, we propose to achieve backdoor attack without modifying the victim model.
Our insight is to inject backdoor logic by attaching a constant input patch, which is feasible since many vision applications have an unchanged foreground/background.
Such an attack is dangerous because (i) it is difficult for model developers to avoid such an attack since the attack happens after the model is securely deployed and (ii) attackers can flexibly control the backdoor logic to implement practical attacks.

The idea of backdooring deep neural networks with an input patch is closely related to adversarial patch attacks \cite{badnet,APbase}, which have been extensively studied in the literature.
However, adversarial patch attacks aim to directly produce a wrong prediction if a carefully-designed patch is presented in the input. 
Instead, our goal is to inject a hidden backdoor logic with a constant patch in the foreground or background.
Our method is a novel connection between the backdoor and adversarial patch attacks.

Our approach includes two main techniques.
First, we adopt a distillation-style training method to generate the backdoor patch without labeled training data. Specifically, we design a training objective that jointly maximizes the patch stealthiness (\ie mimicing the benign model behavior on normal inputs) and the attack effectiveness (\ie producing misbehavior on trigger conditions). 

Second, to enhance the attack effectiveness in the physical world, we propose to model digital-physical visual shift with differentiable transformations (including a shape transformation and a color transformation), so that the digitally-trained backdoor patch can be directly adopted in the physical world.

To evaluate our approach, we perform experiments on three datasets (CIFAR10 \cite{cifar10}, Imagenette \cite{imagenette}, Caltech101 \cite{caltech101}) and three models (VGG\cite{vgg}, ResNet\cite{resnet}, MobileNet\cite{mobilenetv2}). 
The results demonstrate that our attack is robust under different situations with a high attack success rate of 93\% to 99\%.
Meanwhile, our attack is stealthy, since the backdoor patch does not affect the benign accuracy of the victim model, and can hardly be detected with out-of-distribution (OOD) detectors.
We also show that our attack is effective at different levels of over-parameterization by testing it with different pruning ratios (0\%, 30\%, 60\%, 90\%).
By deploying the attack to the physical world, we demonstrate the feasibility of our attack in real-world scenarios.
 
This paper has the following research contributions:
\begin{itemize}
    \item To the best of our knowledge, this is the first backdoor attack against neural networks that does not require any modification on the victim models.
    \item We design a training scheme for the attack, which can generate an effective backdoor patch efficiently with minimal data requirements.
    \item We introduce a digital-physical transformation modeling method that can improve the attack effectiveness in the real-world deployment.
    \item We conduct thorough evaluations of the effectiveness and anti-detection abilities of our attack.
\end{itemize}
The source code is at \url{https://github.com/XaiverYuan/PatchBackdoor}
\begin{figure*}
    \centering
    \includegraphics[width=16cm]{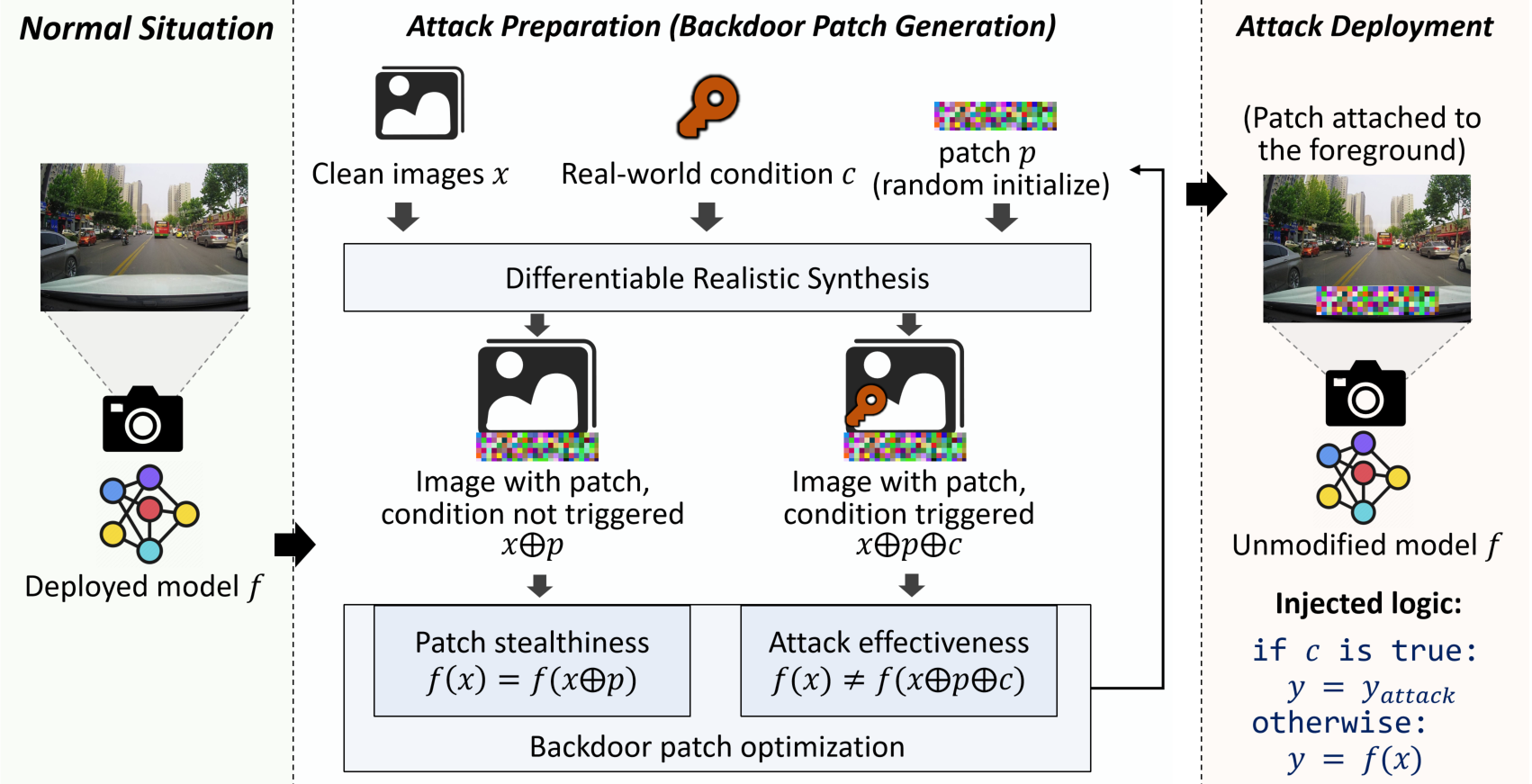}
	\vspace{-0.2cm}
    \caption{The workflow of \sys attack.}
	\label{fig:approach}
\end{figure*}

\section{Background and Motivation}

\subsection{Backdoor Attack against DNN}
\label{section:backdoor_attack}

A deep neural network (DNN) can be viewed as a function $f$ that maps an input $x$ to a prediction $y$.
Typically, a DNN is trained to maximize the following probability:
\begin{equation}
    P(f(x)=\hat{y})
    \label{equation:normalDNN}
\end{equation}
where $f$ is the DNN, $x$ is the image to be classified, $\hat{y}$ is the ground truth label for the corresponding input $x$.

The backdoor attack against a DNN aims to inject a hidden logic into the model, so that the model behaves normally on clean input data, while producing wrong predictions on certain conditions (\eg the input image contains an attacker-controlled trigger).
Formally, the objective of backdoor attacker is to turn the victim model $f$ to a model $f'$ that maximizes following two probabilities:
\begin{equation}
    P(f'(x)=\hat{y}),\ \ P(f'(x\oplus trigger)=y_{t})
    \label{equation:backdoor}
\end{equation}
where $x\oplus trigger$ is the image with the trigger, $y_t$ is the target label for corresponding $x$.

Backdoor attack is difficult to defend against since the attacker-controlled trigger can be arbitrary and unknown to the users.
For example, the backdoor trigger could be invisible to humans if the attackers limit the perturbation boundary \cite{purturbation}. The trigger could also be as small as only one pixel \cite{onePixel}. 
Defending against backdoor attacks typically requires analyzing the poisoned datasets \cite{AC} and/or tuning/retraining the victim model \cite{finePruning}.


A significant limitation of existing backdoor attack approaches is the need to modify the victim model. Specifically, to inject a backdoor, the attacker needs to insert malicious data samples into the training dataset or alter the model.
Since the training data and the model are usually properly protected by the developers, the applicable scenarios of existing backdoor attacks are limited.

Therefore, we are motivated to investigate whether it is feasible to achieve backdoor attack without changing the model and the training datasets. If so, it could pose a significant threat to the security-critical deep learning applications deployed in the real world, since (i) the attacker can flexibly trigger the misbehavior of the model with arbitrary objects and (ii) the attack can be easily achieved after the model is deployed.

\subsection{Opportunity: Backdoor as a Patch}

\textbf{Static foreground/background can be an attack surface.} Many vision applications are deployed to smart cameras with unchanged static foregrounds and/or backgrounds. For example, smart cameras could be deployed in airports to check if terrorists pass by \cite{DNNAirport}. Security cameras are also deployed in military reconnaissance \cite{DNNMilitary}.

Such static foregrounds/backgrounds provide perfect attack surfaces for attaching malicious patches.
It is much easier for an attacker to modify the content in the static foregrounds/backgrounds (which may belong to public spaces) than changing the models that are usually securely protected in the users' private devices. 
The content in the static foreground/background is fed into the model as a part of input, which can interact with the varying content in the input to generate different behaviors.

\textbf{Adversarial patch attack.}
The idea of achieveing attack by controlling a portion of input is not new. Prior research has found that DNNs are vulnerable to adversarial patches, \ie a carefully-designed input patch attached to the input image. 
Such an attack could work because DNNs are known to have redundant neurons, which creates unnecessary logic. Adversarial patches could take advantage of this and activate those redundant neurons to misdirect the prediction. Previous researchers also found that the adversarial patch is a way to draw the attention of the model. 
Therefore, it might be possible to interfere with the \textit{decision logic} of a neural network with an input patch.

Although the adversarial patch attack has demonstrated the ability to change the prediction results with a small patch, it is not practical in the real world because the patch is usually a special image generated by training, which limits the flexibility of controlling trigger conditions.

\textbf{Our idea: Injecting backdoor logic with a patch.}
\label{section:ourIdea}
Instead of directly producing misbehavior with an adversarial patch, our idea is to inject backdoor logic with a patch, where the logic can be controlled by the attacker.

Such an attack has two main advantages.
First, unlike other backdoor attacks that need to change the model, our attack does not require modification on the model and the training data. Second, the trigger conditions of the attack can be flexibly configured, \eg an arbitrary object exists in the camera view, or the environment in under certain lighting conditions.

Meanwhile, implementing the conditional patch attack involve several challenges.
\begin{itemize}
    \item A backdoor attack needs to fulfill two requirements at the same time. First, the attack needs to remain deactivated when the image is normal. Second, the attack needs to be stable when the input image satisfies the trigger condition. Squeezing the logic into a small patch in the input is non-trivial.

    \item It is hard for attackers to obtain the labeled data used for training the victim model. How to generate the backdoor patch with few or no labeled training data is challenging.
    
    \item To be effective in the real applications, the input patch containing the backdoor logic need to be functional in the physical world.

\end{itemize}

\section{Our Approach: \sys}

\subsection{Overview}

\begin{table}[]
    \centering
    \caption{The difference between existing attacks and ours.}
    \vspace{-0.3cm}
    \resizebox{\linewidth}{!}{
    \begin{tabular}{c c c c}
            \toprule
            Method & \makecell[c]{No Model \\ Modification} & \makecell[c]{ Arbitrary\\ Trigger }& \makecell[c]{Physical-world \\Feasibility} \\ \midrule
            Backdoor Attack & False & True & True \\  \hline
            Adversarial Patch & True &False&True\\  \hline
            \makecell[c]{Adversarial Perturbation} & True & False & False \\ \hline
            \sys(ours) &True &True &True \\ \bottomrule
            
    \end{tabular}
    }
    \label{table:compare}
\end{table}

\textbf{Threat Model.}
Suppose there is a deep learning-based vision model deployed in the physical world. The model takes the image captured by a camera as the input, and the camera view contains a constant foreground or background (\eg a wall or a car hood near the camera).
The attacker can inject a backdoor logic (\ie letting the model behavior as usual in most of the time, while producing wrong predictions if the input image satisfies an attacker-defined condition) by attaching a patch onto the constant foreground/background in the camera view.
We assume the attacker has read access to the victim model, so that it can use the model weights to do backward propagation. However, it is impossible for the attacker to modify the model weights.
Moreover, unlike most backdoor attack approaches, we assume that the attacker has no read or write access to the original training data.

The overview of our approach is shown in Figure~\ref{fig:approach}. 
The attack is achieved by training an input patch $p$, namely \emph{backdoor patch}.

In the attack preparation stage, we first obtain the deployed model from the victim application.
Then we choose a real-world condition that could be activated conveniently as the target attacking condition.
We also need to obtain a set of clean images that are normal input images of the victim model. Such an image set should be easy to obtain because the working environment of the victim model is already known.
The backdoor patch is randomly initialized at the beginning.

Based on the clean images, the selected trigger condition, and the randomly-initialized backdoor patch, we can synthesize two types of images. The first is the normal images that have the backdoor patch attached but the trigger condition not satisfied. The second is the target images that contain the backdoor patch and satisfy the trigger condition.
The backdoor patch is iteratively optimized to let the normal images produce normal predictions (so that the patch looks harmless), while letting the target images produce wrong predictions expected by the attacker.

After the backdoor patch is trained, it is deployed to the victim application, by attaching the patch onto a surface in the camera view. In this way, the victim model is unmodified, but its decision logic is altered by the attacker.

\subsection{Backdoor Patch Training}

In Equation~\ref{equation:backdoor}, the objective of our attack is almost the same as the backdoor attack. However, instead of modifying the victim model $f$ to create a new model $f'$, we leave the model $f$ unchanged. $x \in X$ represents a normal image in a dataset $X$. Our patch is denoted as $p$, and $x \oplus p$ means to attach the patch $p$ to the image $x$. 

If an input image satisfies the trigger condition $c$ (\eg a trigger object is present, the lighting is in a specific condition, etc.), we represent it as $x\oplus c.$

Therefore, our attack aims to maximize two probabilities:
\begin{equation}
P(f(x\oplus p)=\hat{y}),\ \ P(f(x\oplus p \oplus c)=y_{target}) \nonumber
\label{equation:cap}
\end{equation}

To meet the above objective, we define the following losses:
\begin{equation}
    L_{clean}(p) = \sum_{x \in X} L(f(x\oplus p), \hat{y})
    \label{equation:loss:stealthiness}
\end{equation}
\begin{equation}
    L_{attack}(p) = \sum_{x \in X} L(f(x\oplus p \oplus c), y_{target})
    \label{equation:loss:effectiveness}
\end{equation}
where $L_{clean}$ is the loss function to encourage patch stealthiness, \ie the normal input with the patch should produce the correct prediction. $L_{attack}$ is the loss to encourage backdoor attack effectiveness, \ie the input image with the patch that satisfies the trigger condition should produce the attacker-specified wrong prediction.

The optimal backdoor patch $\hat{p}$ can be found by minimizing the both losses:
\begin{equation}
    \hat{p} = \underset{p}{argmin}\left(\alpha L_{clean}(p) + (1 - \alpha) L_{attack}(p)\right)
    \label{equation:p_objective}
\end{equation}
where $\alpha$ is a hyperparameter to balance the clean accuracy and attack success rate. If $\alpha$ is closer to one, then the patch will focus more on being stealthy. If $\alpha$ is closer to zero, the patch focuses more on attacking. Attackers could train a standard adversarial patch if we set our $\alpha$ to zero. In practice, setting $\alpha$ to 0.5 is usually a good choice in most cases. However, sometimes $\alpha$ is not easy to determine depending on many factors, including the backdoor patch size, the trigger size, and the image resolution.
In that case, we could adjust the hyperparameter $\alpha$ during training to balance the two losses. This is especially helpful in some cases (\eg when training on small images), where the attack effectiveness loss $L_{attack}$ and stealthiness loss $L_{clean}$ change at significantly different speeds. Using $\alpha$ can balance the two losses and lead to better tradeoff between the clean accuracy and the attack success rate.

As we mentioned before, one challenge in generating the backdoor patch is the lack of labeled data. Thus, the ground-truth label $\hat{y}$ in Equation~\ref{equation:loss:stealthiness} is usually unknown. We borrow the idea of model distillation to solve this challenge - We can consider the original model with the constant patch as a new model $f'(x) = f(x\oplus p)$, and the pixels in the constant patch $p$ are the parameters of the new model $f'$. Then, $f'$ can be trained by distilling knowledge from $f$, \ie Equation~\ref{equation:loss:stealthiness} can be written as:
\begin{equation}
    L_{clean}(p) = \sum_{x \in X} L(f(x\oplus p), f(x))
    \label{equation:loss:stealthiness2}
\end{equation}
In this way, the attacker only needs an unlabeled dataset of clean images, which is easy to obtain, and the trained patch can mimic the behavior of the original model, improving the patch stealthiness.

    
    

\subsection{Digital-Physical Transformation}
\label{section:transformation}

We have so far described how to train the backdoor patch in the digital world. However, in practice, the patch should be attached to a surface in the physical world to conduct the attack. The digitally-trained backdoor patch may become invalid due to the difference between the digital and physical worlds.
Thus, we need to take the digital-physical gap into the consideration when training the patch.

Our key idea is to model the digital-physical gap with a differentiable transformation, and optimizing the backdoor patch using this transformation.

\textbf{Patch Transformation Modeling}. A digital-physical transformation can be separated into two parts, including shape transformation and color transformation.

\begin{figure}
    \centering
    \includegraphics[width=7.5cm]{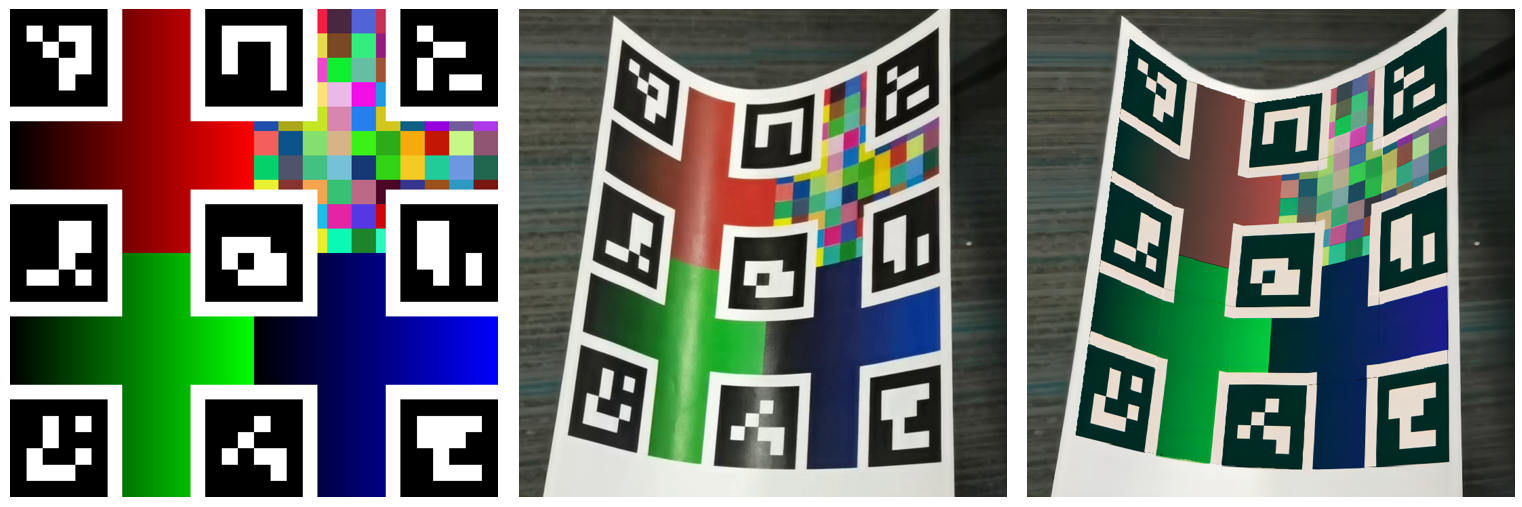}
    \vspace{-0.3cm}
    \caption{The calibration board used for physical-world transformation modeling, including the digital calibration board (left), a photo of the calibration board in the physical world (center), a digital version of the calibration board generated with our differentiable transformation (right).}
    \label{fig:calibration}
\end{figure} 

Both the transformations are captured with a carefully-designed calibration board, as shown in Figure~\ref{fig:calibration}.
When preparing the attack, attackers only need to print the calibration board and put it where they plan to attach the backdoor patch. Then they need to take a few photos of the calibration board, indicating how it will look like in the input of the victim model.
The goal of patch transformation modeling is to find a differentiable function that maps the digital calibration board to its physical-world counterpart.

A regular shape transformation can be implemented as a parameterized warp operation \cite{kornia}, which stretches a square image to fit a quadrilateral. However, the surface to attach the backdoor patch might not be flat, so we propose to split the surface into several smaller surfaces and capture their shape transformations individually.
Our calibration board already contains several ArUco markers that separate the patch to multiple micro-surfaces.
We model the simple shape transformation of each micro-surface and combine them to form the whole shape transformation.

The color transformation is modeled as a lightweight convolutional neural network (CNN). As shown in Figure~\ref{fig:calibration}, our calibration board contains multiple blocks filled with different colors.
By aligning the pixels in the digital calibration board and the physical patch based on the ArUco markers, we can obtain the RGB value mappings between the digital and physical patches. 
We use a single-layer CNN with a 3$\times$3 convolution filter to capture the RGB mapping, and use the MSE loss to minimize the difference between CNN-generated color and the actual color. When the loss converges, the generated CNN is used as the color transformation.

Combining the two transformation modeling techniques, the attacker can flexibly obtain the mapping relation between the digital patch and the physical patch deployed in different environments. In our attack, we consider at least two environments, including the clean environment where the backdoor should remain deactivated and the attacking environment where the patch should cooperate with the trigger to take effect.

Both the shape transformation and the color transformation described above are differentiable, so it is possible to train the patch with backpropagation.
When we want to train a patch that is applicable in the physical world, we simply need to apply the transformations before feeding the patched images into the target model.


\begin{table*}[]
    \centering
    \caption{The clean accuracy and the attack success rate on different models and datasets. ACC stands for clean accuracy. ASR stands for attack success rate. P-ratio stands for poisoning ratio. }
    \vspace{-0.3cm}
    \begin{tabular}{c|c|c|c|c|c|c|c|c|c}
        \toprule
        \multirow{2}{*}{ID} & \multirow{2}{*}{Model}& \multirow{2}{*}{Dataset}  & \multirow{2}{*}{Original Acc.}&\multicolumn{2}{c|}{\sys} &\multicolumn{2}{c|}{BadNet P-ratio=5\% }&\multicolumn{2}{c}{BadNet P-ratio=10\% } \\ 
        & & & &ACC & ASR&ACC&ASR&ACC&ASR\\ \midrule
        1	&MobileNet-V2	&CIFAR 10	&93.61\%	&83.41\%	&95.91\%	&90.45\%	&91.55\%	&89.08\%	&96.00\%\\  \hline 
        2	&ResNet50	&CIFAR 10	&94.26\%	&84.01\%	&95.53\%	&90.30\%	&91.25\%	&90.99\%	&95.39\%\\  \hline 
        3	&VGG16 bn	&CIFAR 10	&93.65\%	&79.00\%	&93.11\%	&88.62\%	&87.78\%	&88.96\%	&96.69\%\\  \hline 
        4	&MobileNet-V2	&Imagenette	&96.05\%	&94.75\%	&98.98\%	&89.43\%	&93.55\%	&92.84\%	&93.61\%\\  \hline 
        5	&ResNet50	&Imagenette	&97.24\%	&90.24\%	&98.30\%	&90.50\%	&91.18\%	&85.20\%	&95.46\%\\  \hline 
        6	&VGG16 bn	&Imagenette	&95.92\%	&95.95\%	&96.82\%	&86.93\%	&92.33\%	&87.57\%	&95.41\%\\  \hline 
        7	&MobileNet-V2	&Caltech	&89.28\%	&85.08\%	&97.64\%	&89.87\%	&77.83\%	&88.84\%	&89.50\%\\  \hline 
        8	&ResNet50	&Caltech	&92.16\%	&88.78\%	&98.00\%	&91.44\%	&76.48\%	&90.75\%	&88.91\%\\  \hline 
        9	&VGG16 bn	&Caltech	&90.95\%	&80.49\%	&93.28\%	&76.25\%	&77.66\%	&76.35\%	&89.41\%\\   
         \bottomrule
    \end{tabular}
    \label{table:results}
\end{table*}

\section{Evaluation}

\begin{figure}
    \centering
    \includegraphics[width=7cm]{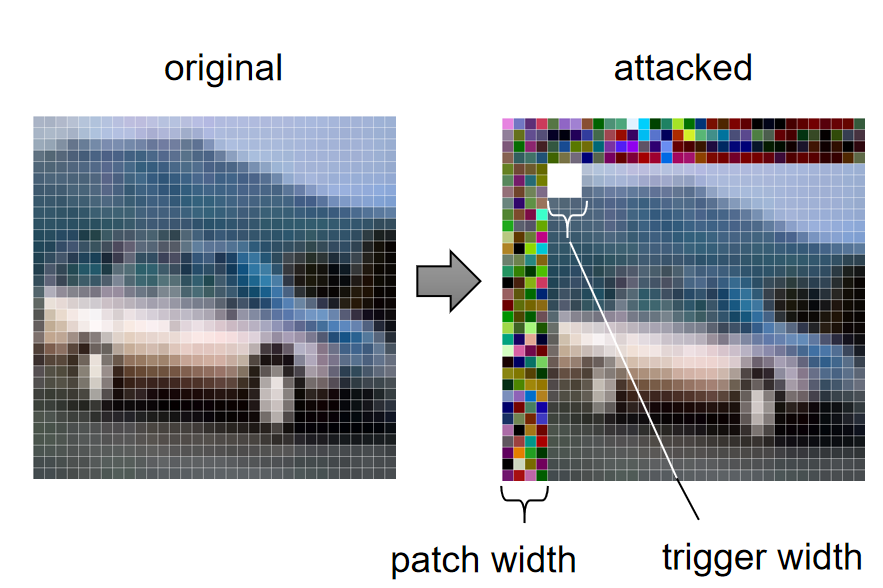}
	\vspace{-0.4cm}
    \caption{An example of how the backdoor patch and backdoor trigger are attached to the image.}
    \label{fig:patchAndTrigger}
\end{figure}

\textbf{Experiment setup.} In most experiments, the backdoor patch is placed as a top-left sidebar in the input image, and the original image is resized to fit into the bottom right corner. 
The backdoor trigger is a white square placed next to the sidebar. The models used in the experiment are all pre-trained on the original datasets. We use patch width and trigger width to describe the size of the patch and the trigger. Figure \ref{fig:patchAndTrigger} illustrates an original image and its corresponding attacked image, as well as the definitions of patch width and trigger width. 
The two important metrics in our experiments are clean accuracy (ACC for short), \ie the accuracy of the model on the normal dataset after the backdoor patch is applied, and attack success rate (ASR for short), \ie the ratio of misclassified images among all images with the backdoor trigger presented.

The experiments are conducted on three CNN models including VGG \cite{vgg}, ResNet \cite{resnet}, and MobileNet-V2 \cite{mobilenetv2} and three datasets including CIFAR-10 \cite{cifar10}, Imagenette \cite{imagenette}, and Caltech \cite{caltech101}.

\subsection{Attack Effectiveness}
\label{sec:influence_of_sizes}

We first evaluate the effectiveness of our attack by measuring the attack success rate and clean accuracy on the common models and datasets. 
Since the three datasets have different sizes, the patch widths and trigger widths vary.
Specifically, the patch width and trigger width are 7 and 3 respectively on CIFAR-10 and 36 and 38 on Imagenette and Caltech.

The results are shown in Table \ref{table:results}. \sys achieves a high attack success rate (93\%-99\%) while maintaining a reasonable clean accuracy. The high attack success rate demonstrates the vulnerability of target models to our attack, while the clean accuracy illustrates that the backdoor patches do not substantially affect the performance of models on normal unperturbed input. 

The accuracy drops for CIFAR-10, Imagenette, and Caltech are around 10\%-15\%, 0\%-7\%, and 4-10\% respectively. The reason why the accuracy drop for CIFAR-10 is higher is probably because the CIFAR-10 dataset contains images of smaller size, resulting in fewer pixels and reduced information capacity in the backdoor patch.

The attack effectiveness results for the three models are close. However, an interesting obversation is that the models with higher original accuracy also produce higher clean accuracy after the backdoor patch is attached. This probably means that the learning abilities of the victim models can be transferred to the backdoor patches when the patches are trained to maximize the stealthiness.

We have also compared the effectiveness of \sys with the classical data poisoning attack (BadNet) on three different datasets. On all datasets, our attack success rates are higher than data poisoning with 1\% poisoning ratio. On datasets with larger image sizes (Imagenette and Caltech), our attack performance is even higher than 10\% data poisoning. However, our attack is slightly less effective than 10\% data poisoning on the CIFAR-10 dataset. This is due to the fact that the reduced image size creates less opportunity for injecting backdoor logic. After all, BadNet is a attack that needs data poisoning and model training.

\begin{figure}
    \centering
    \includegraphics[width=8.5cm]{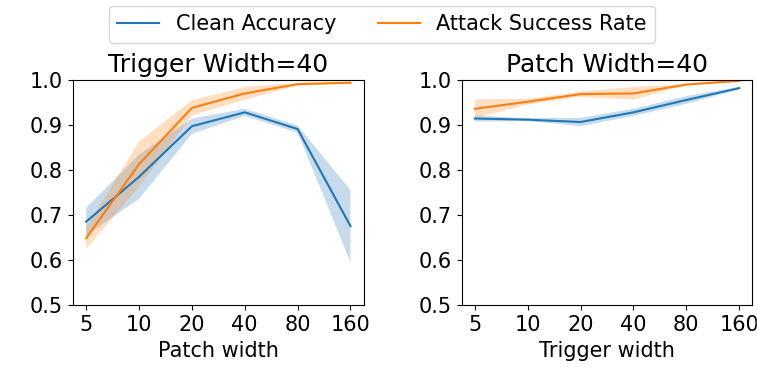}
	\vspace{-0.7cm}
    \caption{The correlation between the attack effectiveness and the sizes of backdoor patches and triggers.}
    \label{fig:sidesize}
\end{figure}

\textbf{Influence of Patch Size and Trigger Size.}
We further analyze the relation between the attack effectiveness and the sizes of backdoor patch and trigger. We select the Imagenette dataset for this evaluation due to its large image size. The victim model is ResNet. All other parameters except the patch width and trigger width held constant at default values. 

As shown in Figure \ref{fig:sidesize}, both the patch size and trigger size influence the clean accuracy and attack success rate. As the patch width increases, both the ACC and ASR increase initially, which is intuitive because the larger patch width represents a larger area for the attacker to manipulate, providing more opportunities to plant backdoor logic. However, as the patch width continues to increase beyond 40, the ASR continues to increase while the ACC starts to drop. This is because the area of the original image becomes too small to carry enough information for classification.

On the contrary, the increase of trigger width constantly leads to the increase of clean accuracy. This may seem counterintuitive since the clean accuracy is measured on clean images that do not contain the trigger. The main reason is that the backdoor patch contained in \sys has two competing objectives - (i) remaining highly accurate under normal conditions, and (ii) producing incorrect predictions when combined with the trigger. \sys achieves both these objectives by optimizing the pixels in the backdoor patch. A larger trigger size makes the latter objective easier to attain, allowing the backdoor patch to focus more on the former objective of improving clean accuracy under normal conditions.

\subsection{Attack Robustness}\label{sec:robust}

We further investigate whether our backdoor patch remains effective after the victim model is modified.
The experiments are all conducted with the Imagenette dataset and ResNet50. Both the patch width and trigger width are set to 40.

\begin{table}[]
    \centering
    \caption{The effectiveness of our attack when trained and evaluated on models with different pruning ratios.}
    \vspace{-0.3cm}
    \resizebox{\linewidth}{!}{
    \begin{tabular}{c|c|c|c|c|c|c|c|c}
        \toprule
        \multirow{2}{*}{\diagbox{Train}{Test}}&\multicolumn{2}{c|}{Prune 0\%} &\multicolumn{2}{c|}{Prune 30\%}&\multicolumn{2}{c|}{Prune 60\%}&\multicolumn{2}{c}{Prune 90\%}\\ 
        &ACC&ASR&ACC&ASR&ACC&ASR&ACC&ASR\\ \midrule
        Prune 0\%&97.5& 98.0& 95.5& 99.3&94.7& 95.8& 93.7& 11.4\\\hline
        Prune 30\%&98.3& 83.2& 97.4& 97.9& 95.6& 94.9& 94.1& 11.9\\\hline
        Prune 60\%&98.5& 13.2& 98.5& 17.8& 96.8& 98.5& 94.7& 10.9\\\hline
        Prune 90\%&98.1& 10.5& 98.0& 10.6& 97.3& 11.3& 94.9& 97.9\\  \hline
        Prune 0\%\&90\% & 96.4 & 98.1 & 94.5 & 99.2 & 94.5 & 94.6 & 94.3 & 97.8\\ \bottomrule
    \end{tabular}
    }
    \label{table:RobustPrune}
\end{table}

\textbf{Robustness against pruning.}
We first consider the case where the victim model is pruned after our attack. Specifically, we consider four pruning ratios of a same model, including 0\% (the original model), 30\%, 60\%, and 90\%. The corresponding model accuracies are 99.46\%, 99.51\%, 99.54\%, and 97.43\%, respectively. The models are pruned and fine-tuned with the global L1 pruning \cite{L1Pruning}.
As shown in Table \ref{table:RobustPrune}, the backdoor patch trained on the original model or the 30\%-pruned models performs well on the 60\%-pruned model. However, the attack trained on models with higher pruning ratios cannot transfer effectively to models with lower pruning ratios. The reason for this is that highly pruned models requiring more fine-tuning, which results in larger modification of the model parameters and subsequently worsens the transferability.
In the last row, the attack is trained on both the original model and the 90\%-pruned model. Surprisingly, the results indicate that the attack remains effective on all other pruned models. This finding suggests that training the patch on multiple models can enhance its robustness.


\textbf{Robustness against fine-tuning.}
Similarly, we test the effectiveness of the patch backdoor after the victim model is fine-tuned. In our study, we initially trained the backdoor patch on the original model, which achieved a clean accuracy of 96.18\% and an attack success rate of 99.13\%. Subsequently, we fine-tuned the model for 40 epochs (accuracy increased from 99.34\% to 99.52\%). Next, we assessed the patch's performance on the finetuned model, revealing a clean accuracy of 95.11\% and an attack success rate of 99.54\%. These results closely resemble the effectiveness observed on the original model, thereby demonstrating the robustness of our attack against normal fine-tuning.
When the model parameters are significantly changed (\eg fine-tuned on different data), the attack trained with the original model may be less effective. In that case, the attacker can re-generate the backdoor patch with the new model, which is quite efficient according to Section~\ref{sec:efficiency}.
%

\textbf{Robustness against distillation.}
We also consider the case when the attack is trained on the original model and applied to a distilled model.
Such robustness is useful when the model parameters are not accessible - attackers can distill a surrogate model using the inference interface and train the attack on the surrogate model.
Specifically, we assume that the attacker is aware of the model structure and has access to a similar distribution of the original dataset. After distilling and training the patch on the surrogate model, we achieve a clean accuracy of 93.63\% and an attack success rate of 99.36\%. When testing the patch on the victim model, we observe a clean accuracy of 92.23\% and an attack success rate of 98.32\%, which are just slightly lower than on the surrogate model.


\subsection{Stealthiness against Detection}

Most defenses against backdoors do not apply to our approach, because they mostly concentrate on identifying or mitigating manipulations made to the training datasets or the models, while our approach does not make any modification to the model architecture, model parameters, training data, and training procedure.
The defenses against adversarial patches also do not apply. Adversarial patch defenses are mostly based on the fact that adversarial patch attacks aim to alter the prediction when the patch is present \cite{mccoyd2020minority,huang2023patchcensor}.
However, the backdoor patches in \sys aim to keep the original predictions, which is a fundamentally different goal as compared with adversarial patches.

However, since the backdoor patch needs to be constantly placed in the camera view, it will alter the data distribution of camera images and may be detected by out-of-distribution (OOD) detectors.
Therefore, we use different OOD detection methods (Baseline\cite{baseline}, FSSD\cite{fssd}, Maha\cite{maha}, ODIN\cite{odin}) to see whether they can distinguish the \sys-modified images with other normal images. 
We use the CIFAR-10 dataset and ResNet model in this experiment, and the patch width and trigger width are 7 and 3 respectively.

We train the OOD detectors with a subset of CIFAR-10 as the in-distribution dataset, and use them to measure the OOD degrees of different out-of-distribution datasets.
The compared OOD datasets include another subset of CIFAR-10 with different classes (CIFAR-10 sep), the CIFAR-100 dataset, and the SVHN dataset. The OOD degree is measured as the AUROC metric, and a higher AUROC means that the dataset is more easily detected as OOD.

The results are shown in Table \ref{table:ood}. We can see that the SVHN dataset can be easily detected as OOD with high AUROC scores, while the AUROC scores for the CIFAR-10 subset and CIFAR-100 are much lower. It is an intuitive result since SVHN is indeed more distributionally different with CIFAR-10 than the other two datasets.
The dataset with \sys (\ie CIFAR-10 images with the backdoor patch attached) yields much lower AUROC scores than SVHN, and sometimes even lower than CIFAR-10 subset and CIFAR-100. This means that our backdoor patches are not easy to detect with common OOD detectors.

\begin{table}[]
    \centering
    \caption{The AUROC scores computed by different Out-Of-Detection detectors for different datasets. The in-distribution dataset is a CIFAR-10 subset, and the compared datasets include another CIFAR-10 subset, CIFAR-100, SVHN, and \sys (CIFAR-10 with a backdoor patch attached). The higher AUROC score means that the compared dataset can be more easily detected as OOD.}
    \resizebox{\linewidth}{!}{
    \begin{tabular}{c|c|c|c|c}
        \toprule
        Data & FSSD\cite{fssd} & Baseline\cite{baseline} & Maha\cite{maha} & ODIN\cite{odin} \\ \midrule
        CIFAR-10 sep & 94.0\% & 59.3\% & 91.4\% &90.4\%\\ \hline
        CIFAR-100 & 74.4\% & 79.8\% &54.8\% &81.5\% \\ \hline
        SVHN & 99.5\% &89.9\% & 99.1\% & 96.6\% \\ \hline
        \sys & 78.3\% &  68.3\% & 68.1\% &69.5\%\\
        
         \bottomrule
    \end{tabular}
    }
    
    \label{table:ood}
\end{table}

\subsection{ACC-ASR Tradeoff under Different Settings}
\begin{figure}
    \centering
    \includegraphics[width=7cm]{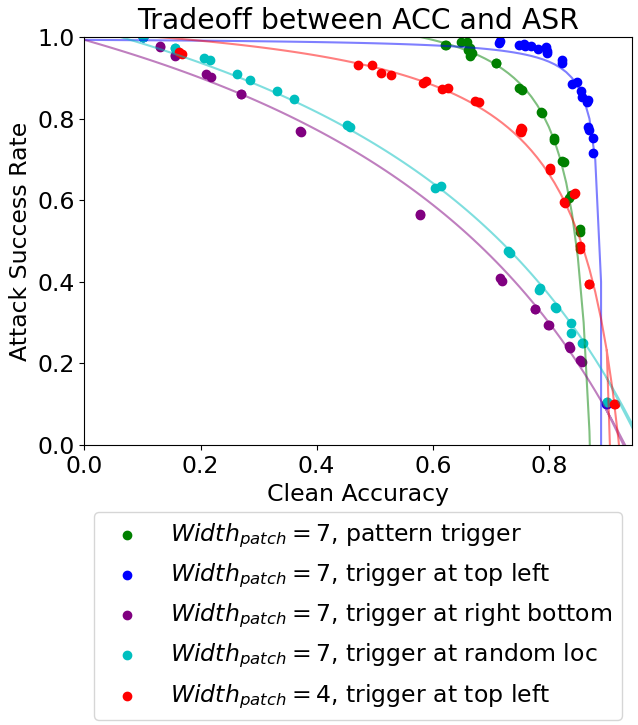}
	\vspace{-0.3cm}
    \caption{The tradeoff between clean accuracy (ACC) and attack success rate (ASR) under different settings.}
    \label{fig:ratio}
\end{figure}

In this subsection, we analyze the tradeoff between clean accuracy (ACC) and attack success rate (ASR) under different settings of \sys.
Our method incorporates a hyperparameter, $\alpha$, that governs the weighting of two competing loss functions during model training. The first loss function optimizes standard classification accuracy on benign examples, while the second aims to maximize the misclassification of adversarial examples to a specific target label. By adjusting $\alpha$, we can precisely control the trade-off between model accuracy on normal inputs and the success rate of the implanted backdoor. 
We use the CIFAR-10 dataset and the ResNet model in this experiment.

In Figure~\ref{fig:ratio}, each point represents an experiment. The experiments are conducted with different settings, including the backdoor patch width, the location of trigger, and the type of trigger.
Overall, we can see that the ACC and ASR exchanges with each other under all settings. This is intuitive as we have mentioned that they are competing goals of our backdoor patch. In cases when the patch size and trigger location are proper (\eg the blue dots), the ACC and ASR can both be high.

If the backdoor patch size is smaller (the red dots), both the ACC and the ASR decrease. The reason has been discussed in Section~\ref{sec:influence_of_sizes}.

The comparison between the blue, clay, and purple curves demonstrates that the attack performance decreases as the trigger appears at positions farther from the patch. This is because that the closer distance between the trigger and the patch makes it easier for combining them to produce misbehavior.
Meanwhile, even when the location is fixed (purple dots), the performance is still worse than that of a randomly positioned trigger (clay dots). This indicates that although the randomness of trigger location may have some impact on the attack performance, the distance between the patch and the trigger is a more influential factor.

We also consider the cases where the trigger is an image pattern instead of a patch.
The green curve means that the trigger pattern is a brightness shift within the image. We can see that \sys can also successfully perform the attack, although the ACC-ASR tradeoff is slightly worse than the blue curve. This demonstrates the flexibility of \sys in customizing trigger conditions.


\subsection{Physical World Feasibility}
\begin{figure}
    \centering
    \includegraphics[width=8.2cm,height=5cm]{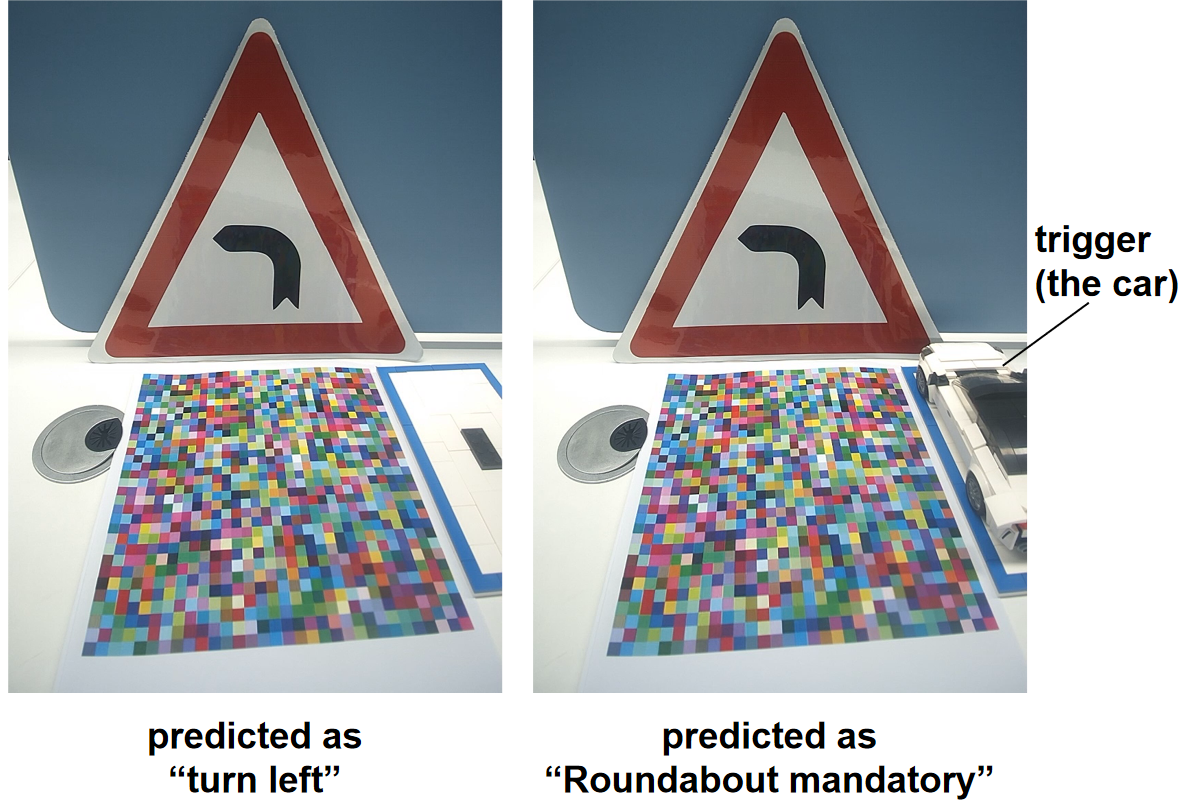}
    \caption{Physical-world feasibility of \sys. The victim model is a traffic sign classifier, the backdoor trigger is the car model on the right.}
    \label{fig:realworldPT}
\end{figure}


In this experiment, we capture images of various traffic signs from different angles and train a customized traffic sign classifier as the victim model. 
Our backdoor patch is trained with the digital-physical transformation (Section~\ref{section:transformation}), printed on a standard A4 paper, and placed below the traffic signs. 
The backdoor trigger is a car model - our attack aims to let the model misclassify the traffic sign when the car model is present.

Our attack achieved a clean accuracy of 90.73\% and an attack success rate of 100\% on our self-collected images. These results demonstrate the feasibility of our attack in the physical world. The high clean accuracy and attack success rate demonstrate that both the stealthiness and attack effectiveness of the generated backdoor patch. The backdoor logic of \sys is robust enough against real-world transformations.

\subsection{Efficiency}
\label{sec:efficiency}
\begin{figure}
    \centering
    \includegraphics[width=6cm]{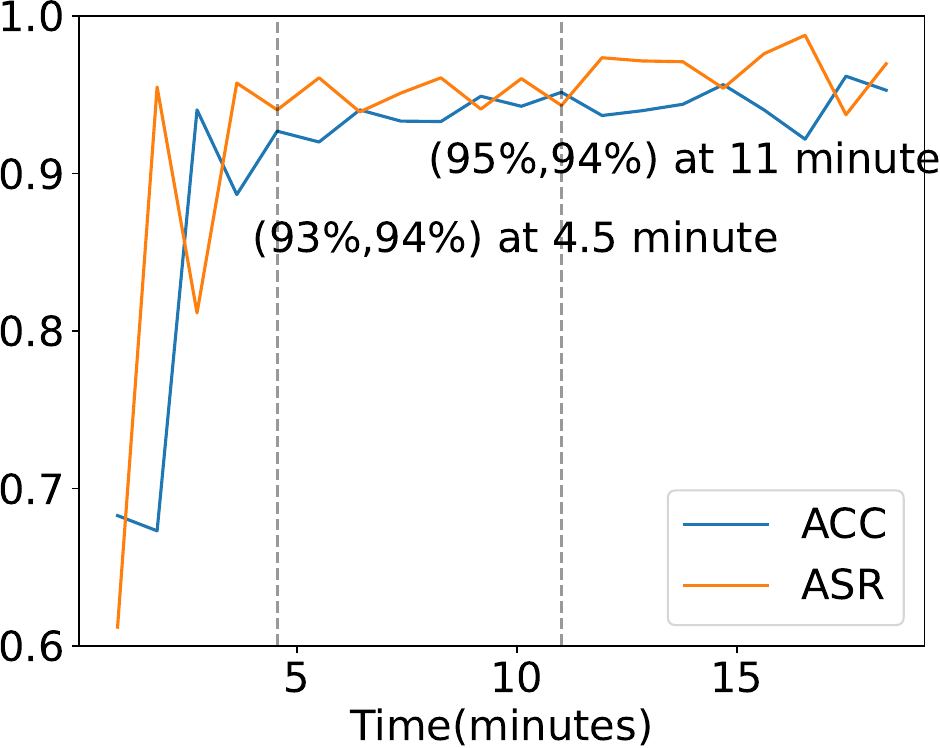}
    \caption{The attacked effectiveness achieved by training with different periods of time.}
    \label{fig:time}
\end{figure}

We evaluated the training efficiency of our attack on a Linux desktop with a NVIDIA GTX 3090 GPU. The Imagenette dataset and the ResNet-50 model were employed. Both the patch width and trigger width were set to 40.

As shown in Figure \ref{fig:time}, the patch training process was efficient. Specifically, the patch achieved a clean accuracy of 92.69\% and an attack success rate of 94.04\% within 5 minutes. The clean accuracy and the attack success rate further increased to 95.15\% and 94.31\% at around 11 minutes. 

\section{Related Work}

\textbf{Adversarial Patch Attack} modifies the pixels within a local region of the image to induce model misclassification. Brown et al. \cite{APbase} first applied a universal and physically achievable patch on victim objects. LaVAN \cite{lavan} and adversarial QR patch \cite{APQRCode} were proposed to improve patch stealthiness. Some other approaches \cite{perceptual} attempt to use different methods like GAN to generate the adversarial patch.
Adversarial reprogramming \cite{elsayed2018reprogramming} discussed the idea of repurposing a neural network with a large adversarial patch, which is the closest to ours, but it was not designed for backdoor attacks, and its large digital patches are unlikely to be feasible in the physical world.
Defenses against adversarial patches are mostly based on saliency map \cite{hayes2018visible, naseer2019local}, adversarial training \cite{gittings2020vax, rao2020adversarial}, small receptive field \cite{xiang2021patchguard, xu2023patchzero}, certification \cite{mccoyd2020minority,huang2023patchcensor}, etc.
Our attack also uses the image patch to pose the threat, while our goal (injecting backdoor logic) is fundamentally different from standard adversarial patch attacks.

\textbf{Backdoor Attack} is aimed at embedding hidden backdoors activated by specific triggers into the model. Existing backdoor attacks can be roughly categorized into poisoning-based approaches \cite{qi2023revisiting, li2021anti} and model editing-based approaches \cite{li2021deeppayload}. The data poisoning methods modify the training data to misled the model to classify certain objects as attacker-specified labels. Various efforts have made poison data more concealable \cite{hu2022badhash}. In model editing approaches, attackers focus on modifying the model parameters or injecting extra malicious modules \cite{doan2021lira}.

To defend against backdoor attacks, most approaches aim to prevent or detect data poisoning \cite{gao2023backdoor, javaheripi2020cleann}, or removing the injected backdoors from the model \cite{wang2019neural, xu2020defending}. 
Our method is also a backdoor attack, but the threat model is fundamentally different - we do not require any modification to the training data and victim model.

\section{Conclusion}

We introduce a backdoor attack against DNN models that injects backdoor logic by attaching a patch in the camera view instead of modifying the training procedure or the model. Experiments have demonstrated the effectiveness of the attack and the feasibility in the physical world. Our work suggests that, besides the training data and the model, the constant camera foreground/background may be an important attack surface in edge AI systems.

\begin{acks}
    This work is supported by the National Natural Science Foundation of China (Grant No. 62272261).
\end{acks}

\bibliographystyle{ACM-Reference-Format}
\bibliography{reference}
\end{document}